# Nonlinear Intensity, Scale and Rotation Invariant Matching for Multimodal Images

Zhongli Fan, Li Zhang, Yuxuan Liu

**Abstract**—We present an effective method for the matching of multimodal images. Accurate image matching is the basis of various applications, such as image registration and structure from motion. Conventional matching methods fail when handling noisy multimodal image pairs with severe scale change, rotation, and nonlinear intensity distortion (NID). Toward this need, we introduce an image pyramid strategy to tackle scale change. We put forward an accurate primary orientation estimation approach to reduce the effect of image rotation at any angle. We utilize multi-scale and multi-orientation image filtering results and a feature-to-template matching scheme to ensure effective and accurate matching under large NID. Integrating these improvements significantly increases noise, scale, rotation, and NID invariant capability. Our experimental results confirm the excellent ability to achieve high-quality matches across various multimodal images. The proposed method outperforms the mainstream multimodal image matching methods in qualitative and quantitative evaluations. Our implementation is available at https://github.com/Zhongli-Fan/NISR.

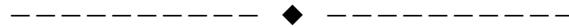

## 1 INTRODUCTION

IMAGE is one of the most widely applied data forms in many areas [1], [2], [3], [4]. Various modalities of images have been developed nowadays, as displayed in Fig. 1. Each modality of images encodes one aspect of information and has its limitations. The fusion of different modalities of images is conducive to the comprehensive utilization of their advantages. As a prerequisite, image matching is crucial to many image fusion applications, such as image registration [5], structure from motion [6], visual simultaneous localization and mapping [7], and so on. Even though the task of image matching has been researched for decades [8], [9], [10], multimodal image matching is still challenging work due to the severe nonlinear intensity distortion (NID) and geometric distortion.

Recent works show that the local image structure recovered from multi-scale and multi-orientation filtering results has good robustness against NID and can be applied to match images with translation [11], [12], [13], [14]. However, very few works consider scale change and image rotation. The scale variance can be modeled with the image pyramid strategy, but the traditional rotation handling strategies become ineffective. Most methods use image gradient or intensity information to estimate a primary orientation for each feature to achieve rotation invariance [15], [16], [17]. However, both image gradient and intensity are unstable under different image modalities due to NID.

Using traditional methods to estimate the primary orientation will cause the wrong estimation and leads to matching failure. Moreover, these methods only use the orientation index information and ignore the complete multi-orientation filtering amplitudes with richer information.

In this paper, we put forward a nonlinear intensity, scale, and rotation invariant method (NISR) for multimodal image matching, which treats these complex factors explicitly. We show that the local structures in the orientation index map are similar under different rotation angles even though the index values change (Fig. 4). Based on this principle, we develop an accurate and robust primary orientation estimation method that analyzes the distribution of the index values, calculates a weighted centroid point, and takes the direction from the feature point to the calculated centroid point as the primary orientation. We also construct a template feature with the complete multi-orientation filtering results to further improve the matching performance of the reference image and the resampled sensed image, which is corrected based on a similarity transformation calculated from the matches obtained in feature matching. In summary, our main contributions are:

1) A robust primary orientation estimation method for multimodal images that uses the local similarity of the orientation index map built from multi-orientation and multi-scale filtering results.

2) A feature-to-template matching framework that first uses the index information to increase robustness against image modality variance and then uses the complete multi-orientation filtering results to improve matching reliability.

3) An effective matching method for multimodal images that takes scale change, image rotation, and severe NID into consideration.

We create an image pyramid and detect distinctive and repetitive feature points on phase congruency (PC) maps generated from each layer image (Sec. 3). We construct a

---

- *Zhongli Fan is with the State Key Laboratory of Information Engineering in Surveying, Mapping and Remote Sensing, Wuhan University, Wuhan 430079, China (e-mail: fanzhongli@whu.edu.cn)*
- *Li Zhang and Yuxuan Liu (Corresponding author) are with the Institute of Photogrammetry and Remote Sensing, Chinese Academy of Remote Sensing (CASM), Beijing 100036, China (e-mail:zhangl@casm.ac.cn, yxliu@casm.ac.cn)*

- *The work was supported by the National Natural Science Foundation of China (42201494).*

*Manuscript received*

feature descriptor involving primary orientation estimation and match the feature descriptors in Sec. 4 and 5. We also present a template feature that can rematch the unmatched feature points (Sec. 6). As seen in Fig. 1, our method can obtain more matches with high accuracy across various multimodal images. In Sec. 7, we verify the robustness against image rotation (Fig. 9 and 10) and scale change (Fig. 11 and 12) and present extensive qualitative (Fig. 13~15) and quantitative results (Table 4) on 164 multimodal image pairs, demonstrating superior results against previous work. Additionally, we apply NISR to image registration (Fig. 16) in Sec. 8.

a co-occurrence scale space to extract image edge features and combined Butterworth and Sobel filters to calculate descriptors. For the same purpose, Li et al. [19] adopted a local normalization filter to resist significant NIDs and improved the ORB detector [20] and HOG descriptor [21] for feature matching. These methods use image gradients to estimate primary orientation to handle image rotation, but the gradient information is unstable for multimodal images, significantly decreasing the performance. Based on multi-scale and multi-oriented log-Gabor filters, Aguilera et al. [11] proposed a multi-layer orientation index map and developed a log-Gabor histogram de-

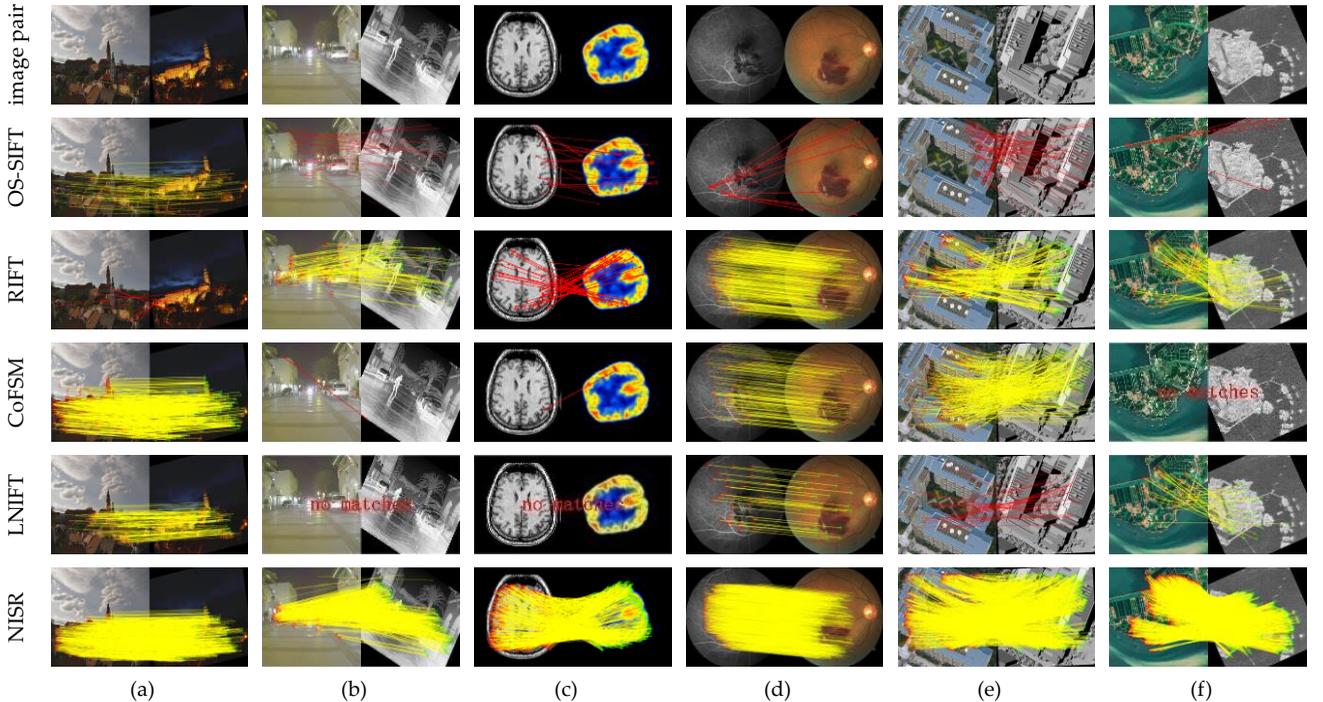

Fig. 1. Comparison of matching results on six typical multimodal image pairs with different modalities. (a) day-night; (b) visible-infrared; (c) MRI-PET; (d) retina-retina (different angiography); (e) optical-LiDAR depth; (f) optical-SAR. It can be seen that NISR significantly improves the matching performance compared with the state-of-the-art methods: OS-SIFT, RIFT, CoFSM, and LNIFT and can be applied to various image pairs with scale change, image rotation, and severe nonlinear intensity distortion.

## 2 RELATED WORK

Multimodal image matching has been an active research topic over the last few years [18]. While the deep learning technique is introduced to this area, this work focuses on designing new handcrafted features that can be effectively applied to real applications. Previous studies have investigated a range of multimodal image matching methods, including feature-based and template-based techniques [9].

**Feature-based methods:** Chen et al. [15] proposed a partial intensity invariant feature descriptor using the average squared gradient and integrated it into the framework of the scale invariant feature transform (SIFT) algorithm [1]. Xiang et al. [16] used multi-scale Sobel and ROEWA filters to calculate image gradients for optical and synthetic aperture radar (SAR) images and conducted image matching with SIFT algorithm. To match multimodal remote sensing images, Yao et al. [17] constructed scriptor to match visible, depth, and long-wave infrared images, obtaining good matching performance and demonstrating good robustness to severe NID. However, the method can only tackle image translation and does not resist image rotation and scale change. Following this thought, Li et al. [14] proposed a radiation-variation insensitive feature transform (RIFT) algorithm. RIFT utilizes the PC model for feature detection and constructs a maximum index map for feature description. By setting an end-to-end annular structure for multiple index maps, RIFT achieves rotation invariance under small image rotation angles. Besides, the rotation performance is unstable on various image modalities, and the efficiency is extremely low. To address this problem, we propose to use the local similarity of the orientation index map to estimate primary orientation, achieving invariance in any rotation angle across various image modalities. Besides, we introduce the scale space to achieve scale invariance.

**Template-based methods:** Maes et al. [22] proposed using mutual information (MI) to match multimodal medical images. Liu et al. [23] embedded local self-similarity with MI for the matching of optical and SAR images. Due to the high computation complexity of MI, only local maximization is applied to match the images in most situations, and no guarantee with respect to global energy is employed. Researchers recently found that image structures are well retained across different modalities. Ye et al. [12] proposed a histogram of the oriented PC method based on the classical HOG algorithm. Xiong et al. [24] proposed the rank-based local self-similarity algorithm for matching optical and SAR images by introducing a rank correlation coefficient. Unlike these methods that build sparse features, Ye et al. [13] proposed the channel features of orientated gradients by changing the sampling pattern of HOG from the sparse grid to pixel-by-pixel. Fan et al. [25] further proposed the angle-weighted oriented gradients algorithm by distributing the gradient magnitude to the two most relevant gradient directions. Similarly, Zhu et al. [26] used an odd-symmetric Gabor filter to construct dense structure features for matching the optical aerial and LiDAR intensity images. Even though the template-based methods have relatively high matching accuracy, they are susceptible to image geometric distortions, including scale change and image rotation. Considering this, we first use matches obtained in the proposed feature matching process to coarsely eliminate the geometric difference between images and then construct dense template features on the corrected images to improve the matching performance. Particularly, our template feature encodes the complete multi-scale and multi-orientation image information, having good robustness against NID involved in multimodal images.

## 3 FEATURE EXTRACTION

Distinctive and repeatable feature point detection is crucial to the success of feature matching. In this study, we first build an image pyramid to tackle scale change. Then, we compute a salient edge map using multi-orientation phase congruency [27] for each layer image in the pyramid and use the FAST detector [28] to locate feature points. The edge map reduces the affection of image intensity and illumination and enhances the image structure information, increasing the robustness against NID. Besides, the method can work well in low-texture and noisy areas.

### 3.1 Scale-space construction

We construct a scale-space by conducting a serial of image subsampling and Gaussian smoothing to achieve scale invariance. Precisely, the scale-space pyramid consists of $n$ octave layers, $a_i$, and $n$ intra-octave layers, $b_i$, $i = \{0,1,...,n-1\}$. $a_0$ corresponds to the original image, and $b_0$ is obtained by subsampling $a_0$ with a scale factor of 1.5, making the size of $b_0$ is two-thirds of $a_0$. The other octave and intra-octave layers can be extracted by progressively half-sampling the layers $a_0$ and $b_0$, respectively. Typically, $n$ can be set as 4. The construction of the scale-space pyramid can be represented as follows:

$$a_i(x,y) = \sum_{-m}^{m}\sum_{-n}^{n} G(m,n) * a_{i-1}(2x+m, 2y+n) \quad (1)$$

$$G(m,n) = \frac{1}{\sqrt{2\pi\sigma^2}} e^{\frac{-(m^2+n^2)}{2\sigma^2}} \quad (2)$$

where $G(m,n)$ is a Gaussian kernel with standard deviation of $\sigma$.

### 3.2 Feature detection

For each layer image in the pyramid, we calculate a weighted moment map derived from its corresponding PC maps and detect feature points on it. For an image $I(x,y)$, we first convolve it with a multi-scale and multi-orientation log-Gabor filter [29]. This process can be described as follows:

$$RI_{so}(x,y) = \mathcal{F}^{-1}(P(u,v) \cdot LGF_{so}(r,\theta)) \quad (3)$$

$$P(u,v) = \mathcal{F}(I(x,y)) \quad (4)$$

$$LG_{so}(r,\theta) = e^{\frac{-(ln(r/f_s))^2}{2 \cdot (ln(\sigma_r))^2}} \cdot e^{\frac{-(\theta-\theta_o)^2}{2\sigma_\theta^2}} \quad (5)$$

where $RI_{so}(x,y)$ denotes the filtering result; $LG_{so}(r,\theta)$ denotes a 2D log-Gabor filter; $s$ and $o$ are the scale and orientation index, respectively; $(r,\theta)$ represents the polar coordinates; $f_s$ and $\theta_o$ represent the filter center frequency at the scale of $s$ and the orientation angle of $o$; $\sigma_r$ and $\sigma_\theta$ denote radial and tangential bandwidths. $\mathcal{F}(\cdot)$ and $\mathcal{F}^{-1}(\cdot)$ represent the Fourier transform and the inverse Fourier transform, respectively.

Then, a PC map can be obtained with (6) based on the amplitude component $A_{so}(x,y)$ and phase component $\phi_{so}(x,y)$ calculated using the real part $REAL_{so}$ and imaginary part $IMAG_{so}$ of $RI_{so}(x,y)$.

$$A_{so}(x,y) = \sqrt{REAL_{so}(x,y)^2 + IMAG_{so}(x,y)^2} \quad (6)$$

$$\phi_{so}(x,y) = tan^{-1}\left(\frac{IMAG_{so}(x,y)}{REAL_{so}(x,y)}\right) \quad (7)$$

$$PC(x,y) = \frac{\sum_s \sum_o w(x,y) \lfloor A_{so}(x,y)\Delta\Phi_{so}(x,y) - T \rfloor}{\sum_s \sum_o A_{so}(x,y) + \varepsilon} \quad (8)$$

$$\Delta\Phi_{so}(x,y) = cos\left(\phi_{so}(x,y) - \overline{\phi}(x,y)\right) - \left|sin\left(\phi_{so}(x,y) - \overline{\phi}(x,y)\right)\right| \quad (9)$$

where $w(x,y)$ is a weighting factor for frequency domain spread; $\lfloor \cdot \rfloor$ is a truncation function, the value is itself when it is positive, and 0 otherwise; $\Delta\Phi_{so}(x,y)$ is a phase deviation function; $T$ is a noise threshold that can be determined from the response of the filter to the image; $\varepsilon$ is a small value to avoid division by 0; $\overline{\phi}(x,y)$ denotes the mean phase component.

Based on the obtained PC map, a maximum moment map $M_{max}$ and a minimum moment map $M_{min}$ can be calculated using (10) and (11).

$$M_{max} = \frac{1}{2}\left(A + C + \sqrt{(A-C)^2 + B^2}\right) \quad (10)$$

$$M_{min} = \frac{1}{2}\left(A + C - \sqrt{(A-C)^2 + B^2}\right) \quad (11)$$

where,

$$A = \sum_o (PC(\theta_o)\cos(\theta_o))^2 \quad (12)$$

$$B = 2 \cdot \sum_o (PC(\theta_o)\cos(\theta_o))(PC(\theta_o)\sin(\theta_o)) \quad (13)$$

$$C = \sum_o (PC(\theta_o)\sin(\theta_o))^2 \quad (14)$$

where $PC(\theta_o)$ is the PC map under the orientation angle $\theta_o$.

According to the moment analysis theory [30], $M_{max}$ encodes feature information, $M_{min}$ encodes edge information, and $M_{min}$ is a strict subset of $M_{max}$. Considering that the edge structures are more stable across multimodal images, we suppress the corner response and strengthen the edge response by integrating $M_{max}$ and $M_{min}$ to weaken the influence of intensity variations, which can be expressed as follows:

$$\mathcal{W} = \tau \cdot M_{max} + (\tau - 1) \cdot M_{min} \quad (15)$$

where $\tau$ denotes the weight coefficient, ranges from 0.5 to 1.

Finally, the FAST detector [28] is used for feature extraction on the weighted moment map. Fig. 2 shows the feature extraction results on a pair of optical-depth images. We can see that the moment map can well keep the image structure regardless of different image modals, and a large amount of repeated feature points are evenly distributed on the two images.

## 4 FEATURE DESCRIPTION

In this section, we introduce a novel feature descriptor, show how to construct it and demonstrate why it is rotation invariant. As displayed in Fig. 3, we first extract an orientation index map from the multi-orientation filtering results. Then, we evaluate a primary orientation for each point by analyzing the distribution of index values in a local circular area. At last, a high-dimensional feature vector is created by arranging the histogram values of multiple sub-regions in a local square patch.

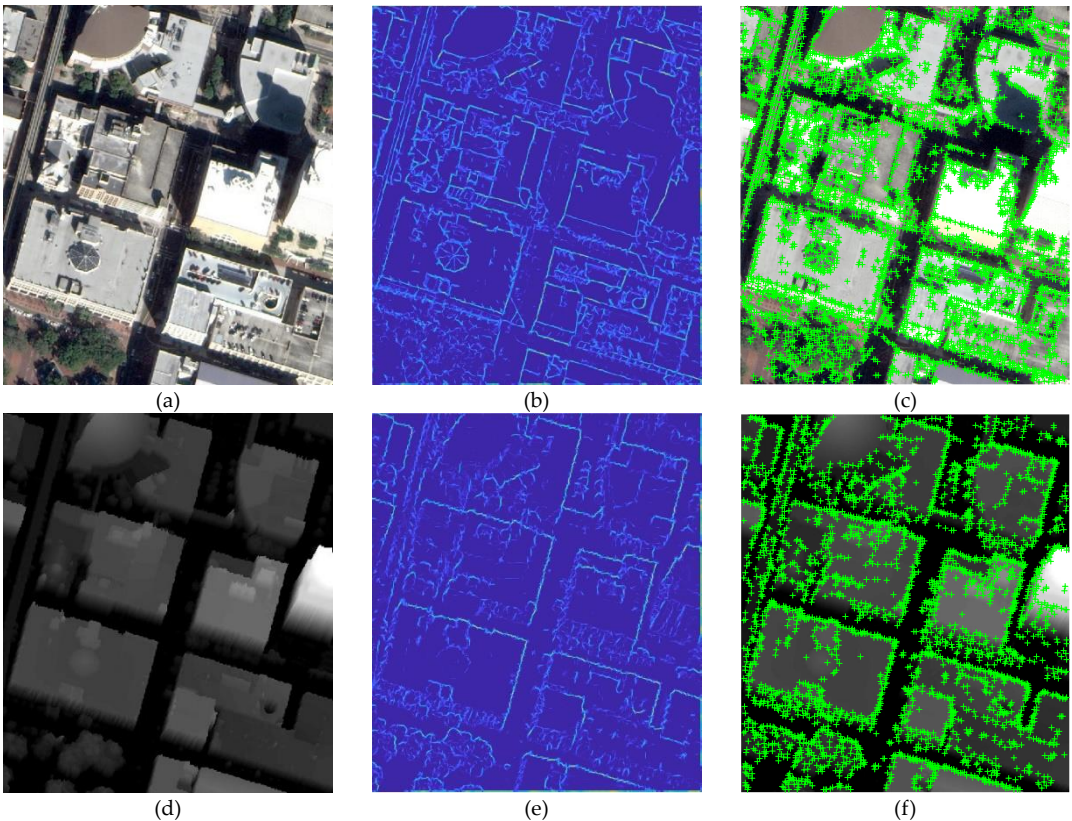

Fig. 2. The feature extraction results on a pair of optical-depth images. (a) and (d) are the original optical and depth images; (b) and (e) are the edge maps calculated by equation 16; (c) and (d) demonstrate the detected feature points.

## 4.1 Orientation index map extraction

Previous studies [11], [14] have proven that the index information of the largest amplitude components in all orientation filtering results is more robust and effective than the amplitude values in describing the feature points for multimodal images. However, they did not give the profound reason behind this principle. Based on detailed experiments and analysis, we found that the reason could be that the largest amplitude components encode image salient structures, and the index information provides a unified measure while the value itself is unstable for multimodal images with large NID. Thus, we first extract an index map from the multi-scale and multi-orientation convolved images. To increase the stability of filtering results in one orientation and improve the efficiency, we add the amplitude components obtained in all scales for each orientation with (16).

$$A_o(x,y) = \sum_{s=1}^{n} A_{so}(x,y) \qquad (16)$$

The index map can be found as follows:

$$Map(x,y) = Index\left(max(A_o(x,y))\right) \qquad (17)$$

where $Index(\cdot)$ and $max(\cdot)$ are used to grab the layer index and find the largest amplitude in all orientations, respectively. Fig. 3 shows the construction process of the index map.

A proper value of $o$ is crucial to the success of image matching. A large value of $o$ will reduce the stability of the following primary orientation estimation and increase computation cost. A small value of $o$ means a large interval between two adjacent orientations of filters, which will decrease the similarity of orientation index maps under small image rotations. Generally, the index value for a pixel has the highest creditability when the rotation angle is an integral multiple of $180°/n$. Simultaneously, when the rotation angle between two images is an integral multiple of $180°/2n$, the index value is most easily biased to the wrong value, which is $180°/n$ when the correct one is $180°/n+1$ or $180°/n+1$ when the correct one is $180°/n$. Toward this, we induce a double index map strategy. Instead of applying one index map extracting from all layers in $A_o(x,y)$, we extract an odd index map using the odd-numbered layers in $A_o(x,y)$, and an even index map using the even-numbered layers in $A_o(x,y)$. The most ambiguous angle for one index map is the most distinguished one on the other, ensuring high performance and decreasing the computation cost. Table 1 gives the performance comparison with/without this strategy on 164 multimodal image pairs described in Sec. 7. We can see that the evaluation metrics presented in Sec. 7 are all significantly improved. Namely, the $SR$ is enhanced by about 14%, the $NM$ is improved by about 280, and the $RMSE$ is improved by about 0.3 pixels with the double index map when $o = 12$.

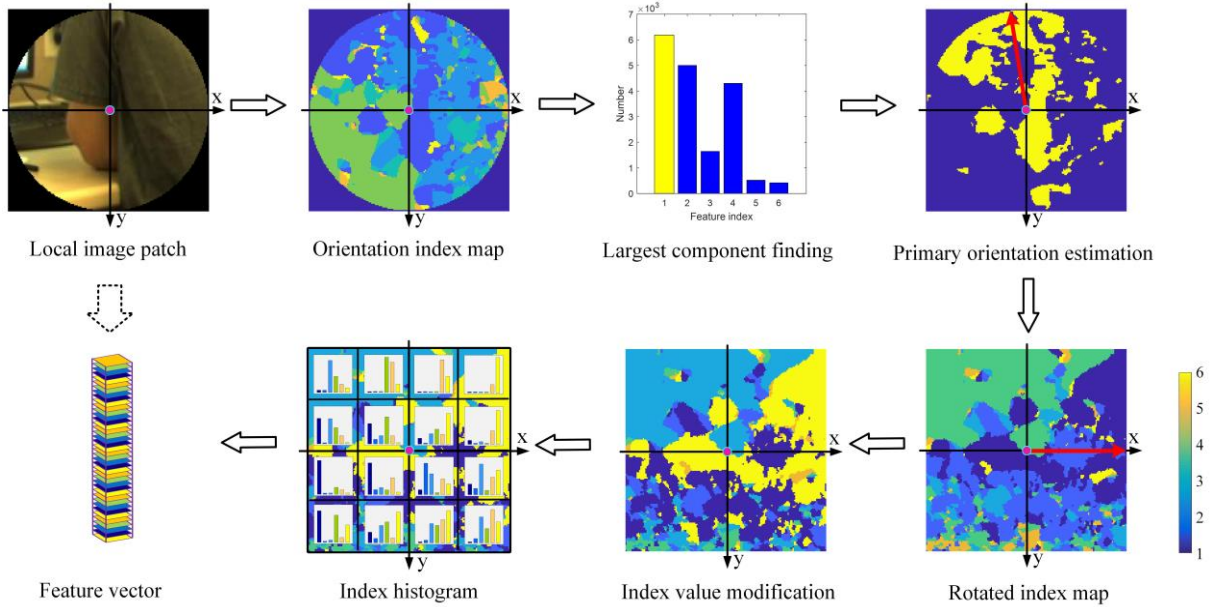

Fig. 3. Feature description.

## 4.2 Primary orientation assignment

The appearance of index maps for different multimodal images without image rotation would be similar (Fig. 4). However, this principle is broken when the images are rotated at different angles, considering that the orientation angles of the log-Gabor filters are fixed. We manually rotate a pair of optical-infrared by different angles, extract an index map at each angle, and demonstrate their visualization results in Fig. 4.

By observing the index maps at different rotation angles, we can see that the local structures are consistent even though the index values change (marked by the red squares in Fig. 4). This is because the general image internal structures of local regions remain unchanged at different rotation angles, so the local general structures on the index map are kept correspondingly. In contrast, the index values become unpredictable, considering that the directions of the multi-orientation filters are fixed when the image rotates at different angles.

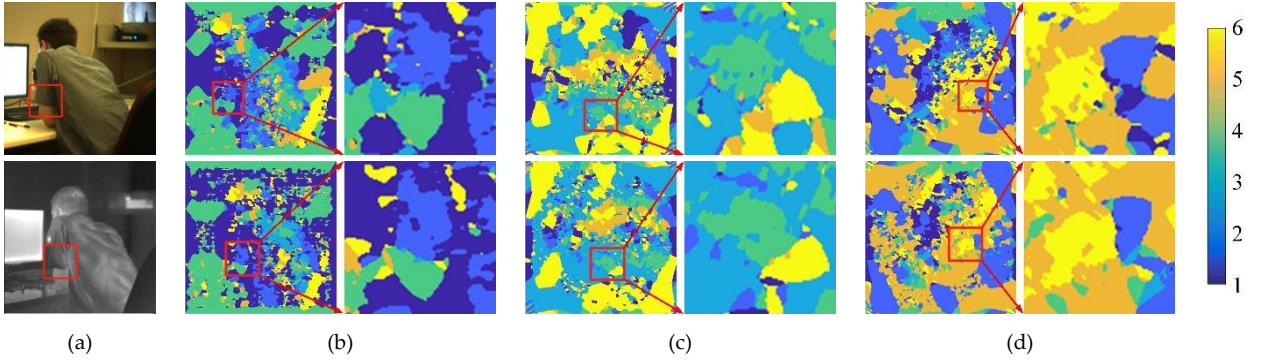

Fig. 4. Appearance of orientation index maps under different rotation angles. (a) optical (up)-infrared (down); (b)~(d) correspond to the maps with enlarged areas marked in red squares at the rotation angles of $0°, 60°, 120°$.

Based on this characteristic, we evaluate the primary orientation for a feature point to recover the similarity of index maps of images under different rotation angles. In detail, a circular area centered at the feature point is first determined on the index map. Then, we count the index value distributions of all the pixels located in the area, find the pixel set whose index value takes the largest percentage, and estimate the coordinates of the centroid point of all pixels. The direction of the line connecting the feature point and the gravity point is taken as the primary direction.

Let P be the point set in the local circular area of p and $p_i$ be a point in P. Taking the coordinates of $p_i$ as the initial coordinates of the centroid point, we gradually update the coordinates by involving the unprocessed pixels in P with (18).

$$G = a + \frac{N_p}{1+N_p} \cdot \overrightarrow{aG} \qquad (18)$$

where G is the updated centroid point after taking the unprocessed point $b$, $a$ is the obtained centroid point after processing $N_p$ points in P

After all the pixels are processed, the coordinates of the centroid point are obtained, and the primary orientation can be further calculated. Considering that the image filters also have orientation, we unify the index of filters in the same direction on different images. As shown in Fig.5, we modify the index values of the pixels whose index values take the largest proportion in the region to $o$ and adjust the index values of the remaining pixels correspondingly, which can be represented as follow:

$$K_i = k_i + (o - k_{mode}), k_i \leq k_{mode} \qquad (19)$$
$$K_i = k_i - (o - k_{mode}), k_i > k_{mode} \qquad (20)$$

where $i$ is an integer and $i \in [1, o]$.

In addition, if the number of pixels for the second primary orientation is larger than 80% of that of the primary orientation, we construct a feature descriptor for the point with the second primary orientation. This strategy can improve the robustness of feature matching. Taking the method that only considers the primary orientation as the baseline, the results on 164 multimodal image pairs (Table 1) demonstrate the effectiveness of the proposed strategy.

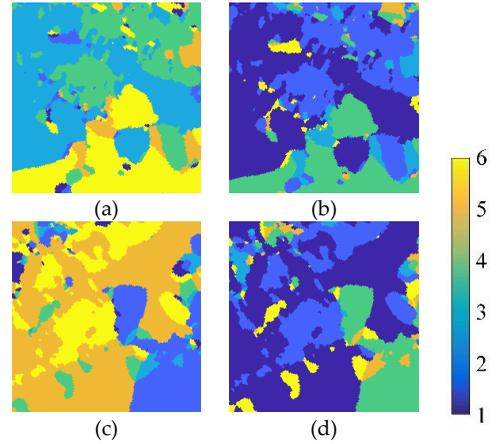

Fig.5. Orientation index value modification. (a) and (c) are the local index maps of the optical image (Fig.4c) and infrared image (Fig. 4d), where the two images are rotated at $60°$ and $120°$, respectively. (b) and (d) are the maps after modification. We can see that index maps become similar after modification.

TABLE 1
RESULTS OF FEATURE MATCHING WITH/-OUT THE PROPOSED STRATEGIES. ORI. DENOTES THE ORIGINAL METHOD. STR.1 DENOTES THE STRATEGY CONSIDERING SECONDARY PRIMARY ORIENTATION; STR.2 DENOTES THE STRATEGY CONSIDERING DOUBLE INDEX MAPS

| Method | NM↑ | RMSE (pixels)↓ | SR (%)↑ |
|---|---|---|---|
| Ori. | 605 | 2.48 | 82.3 |
| Ori.+Str.1 | 723 | 2.39 | 85.4 |
| Ori.+Str.2 | 888 | 2.19 | 96.3 |
| Ori.+Str.1+Str.2 | 1012 | 2.15 | 98.8 |

### 4.3 Feature descriptor construction

The scale, orientation, and location have been assigned to each feature point using the previous operations. Given that a square window on the index map at the given scale has been opened, we divide the square window image into $n \times n$ sub-regions and use a distribution histogram technique for feature vector description. As shown in Fig. 3, we calculated the accumulated values in every value of $o$ in each subregion and connected the statistical results in all subregions to generate the feature descriptor. Moreover, the $o \times n \times n$ dimensional feature vector is normal-ized to reduce the effect of change in illumination.

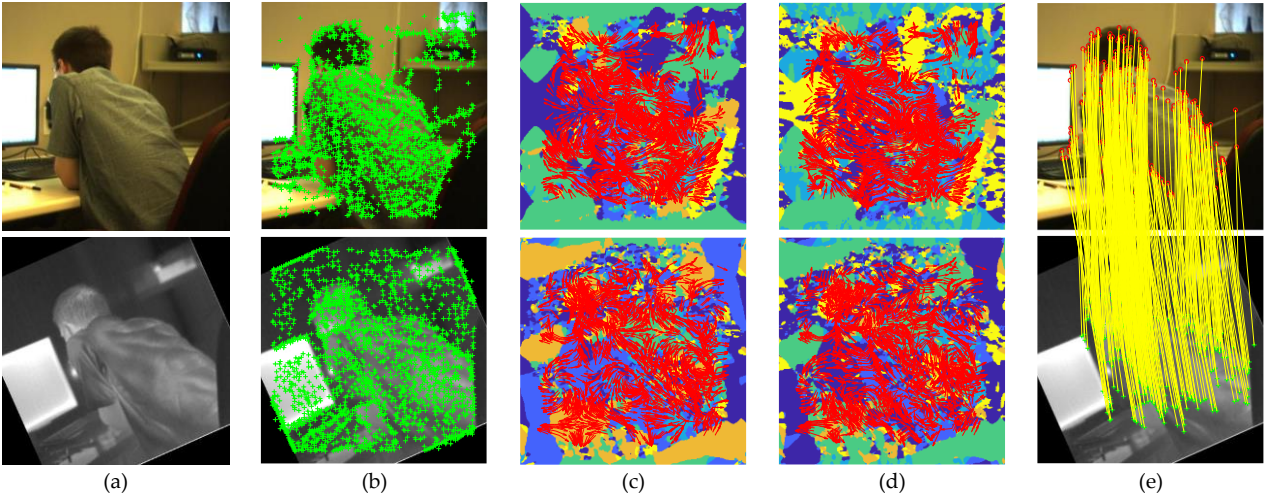

Fig. 6. Results of different stages of our feature matching method. (a) a pair of optical-infrared images with rotation; (b) extracted feature points; (c) and (d) are the two index maps with estimated primary orientation for each feature point, which are indicated with red arrows. (e) established correspondences.

## 5 FEATURE MATCHING

We accomplish feature matching based on the nearest neighbor principle and complete outlier removal using the FSC algorithm [31]. Fig. 6 shows the initial matching results on a pair of optical-infrared images and some intermediate results of this stage. More qualitative and quantitative experimental results will be presented in Sec. 7.

## 6 TEMPLATE MATCHING

Feature matching can successfully match the distinctive features across image modalities, while the less distinctive features are failed to be matched due to local structure change and severe NID. In this section, we rematch the unmatched high-quality feature points using template matching. Note that we utilize the full-orientation filtering results rather than the index information to construct a high-dimensional template feature to improve the matching performance.

First, we calculate a similarity transformation matrix M based on the matches obtained in feature matching and resample the sensed image with M to coarsely eliminate the effect of scale change and image rotation. Then, we use the accumulated amplitude component $A_o(x,y)$ to construct template features. For the reference images, we can directly use the results generated in the previous stage; for the resampled sensed image, we perform a log-Gabor filter on it and calculate $A_o(x,y)$ with (17). Since image rotation between the two images has been almost eliminated, the value of $o$ does not need to be significant to calculate the primary orientation, and a smaller value of $o$ is sufficient for effective matching, as demonstrated in Sec. 7.2. Empirically, $o$ can be set as half of that in the feature matching stage. Finally, the $o$-dimensional template feature $T(x,y,o)$ is constructed by stacking the values of $A_o(x,y)$ in the order of orientation. Moreover, we normalize $T(x,y,o)$ to increase the robustness against illumination change.

To match the template features efficiently, we use phase correlation instead of traditional similarity measures, such as normalized cross-correlation (NCC). Assuming that $T_1(x,y,o)$ and $T_2(x,y,o)$ are template features of the reference image and the resampled sensed image, the geometric relationship between them can be described as

$$T_1(x,y,o) = T_2(x-x_0, y-y_o, o) \qquad (21)$$

where $x_0$ and $y_o$ are the offsets in horizontal direction and vertical direction, respectively.

By performing 3D Fourier transform to $T_1(x,y,o)$ and $T_2(x,y,o)$, we can obtain $FT_1(u,v,w)$ and $FT_2(u,v,w)$. According to the Fourier shift theorem, the correlation between $FT_1(u,v,w)$ and $FT_2(u,v,w)$ can be represented as follow:

$$FT_1(u,v,w) = FT_2(u,v,w)e^{-i(ux_0+vy_0)}\vec{\lambda} \qquad (22)$$

where $\vec{\lambda}$ denotes a 3D unit vector. Then, we can obtain the cross-power spectrum as follow:

$$FT_1(u,v,w)FT_2(u,v,w)^* = e^{-i(ux_0+vy_0)}\vec{\lambda} \qquad (23)$$

where $*$ denotes complex conjugate.

After that, we perform a 3D inverse Fourier transform to the cross-power spectrum, a relation function $\delta(x-x_0, y-y_0)$ is obtained.

$$\mathcal{F}^{-1}(FT_1(u,v,w)FT_2(u,v,w)^*) = \delta(x-x_0, y-y_0)\vec{\lambda} \qquad (24)$$

The optimal matching position can be obtained by searching the largest response value of $\delta(x-x_0, y-y_0)$ in the local image window. Generally, the largest response of the related function will appear at $(x_0, y_0)$.

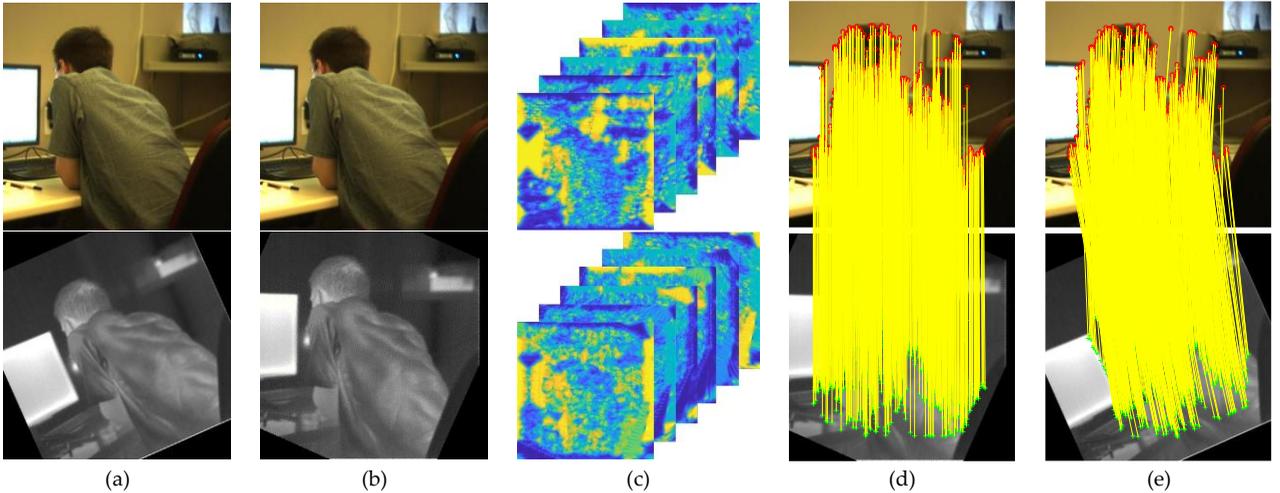

Fig. 7. Results of different stages of our feature matching method. (a) a pair of optical-infrared images with rotation; (b) the original optical image and a simulated infrared image using a similarity matrix estimated by the matches obtained in feature matching; (c) constructed template features; (d) established correspondences on the image pair of (b); (e) established correspondences on the image pair of (a).

After obtaining the matches, the FSC algorithm is adopted for outlier removal. Moreover, the matching points on the resampled sensed image are reprojected back to the coordinate system of the original sensed image. Fig. 7 shows the enhanced matching results on a pair of optical-infrared images. We can see that many more correct matches are obtained on the basis of feature matching, proving the high accuracy of feature matching and the effectiveness of template matching.

## 7 EXPERIMENTAL RESULTS

We first give the general parameter settings for our proposed method, then verify the great invariant against scale change and rotation, and demonstrate the superiority of our method by comparing the qualitative and quantitative results with several state-of-the-art algorithms, OS-SIFT [16], RIFT [14], CoFSM [17], and LNIFT [19] at last. The parameters recommended by the authors for each method are applied. Specifically, we modify the threshold of detecting features for OS-SIFT to 0.001 to obtain more feature points.

Three sets of multimodal images from the fields of computer vision, medicine, and remote sensing are employed, with each set containing six types of challenging multimodal image pairs. Therefore, broad modalities of images are applied, including visible images, infrared images, thermal images, optical images, SAR images, map images, LiDAR depth images, LiDAR intensity images, retina images captured by different angiography techniques, Magnetic-Resonance-Imaging images (MRI), Positron-Emission-computed-Tomography images (PET), Proton-Density-Weighted-Image images (PD), T1-Weighted-Image images (T1), T2-Weighted-Image images (T2), Single-Photon-Emission-Computed-Tomography images (SPECT), Computed-Tomography images (CT), and so on. These images contain significant nonlinear radiometric differences and geometric differences, such as scale, rotation, and translation. A detailed description of the datasets is given in Table 2, and some example image pairs for each set are presented in Fig. 8.

For quantitative evaluation, 10~15 high-precision checkpoints are manually selected for each image pair. For a pair of checkpoints, $(x_1, y_1)$ and $(x_2, y_2)$, a point $(x_1^{'}, y_1^{'})$ can be estimated with a similarity matrix calcu-lated based on the obtained matches. Then, the metrics of the number of putative matches (*NM*), root mean square error (*RMSE*), and success rate (*SR*) can be derived as follows:

$$RMSE = \sqrt{\frac{1}{NM}\sum_{i=1}^{NM}\left[\left(x_2^i - x_1^{i'}\right)^2 + \left(y_2^i - y_1^{i'}\right)^2\right]} \quad (25)$$

$$SR = \frac{N_{success}}{N_{total}} \quad (26)$$

where $N_{success}$ and $N_{total}$ are the number of successfully matched image pairs and the total number of image pairs, rese. *RMSE* encodes the matching accuracy. The lower the *RMSE*, the higher the accuracy. In addition, if *RMSE* is larger than 5 pixels, the result will be deemed as a matching failure. *SR* reflects the robustness and generality of the method for a certain type of multimodal image pair. the higher the *SR*, the better the robustness and generality of the method.

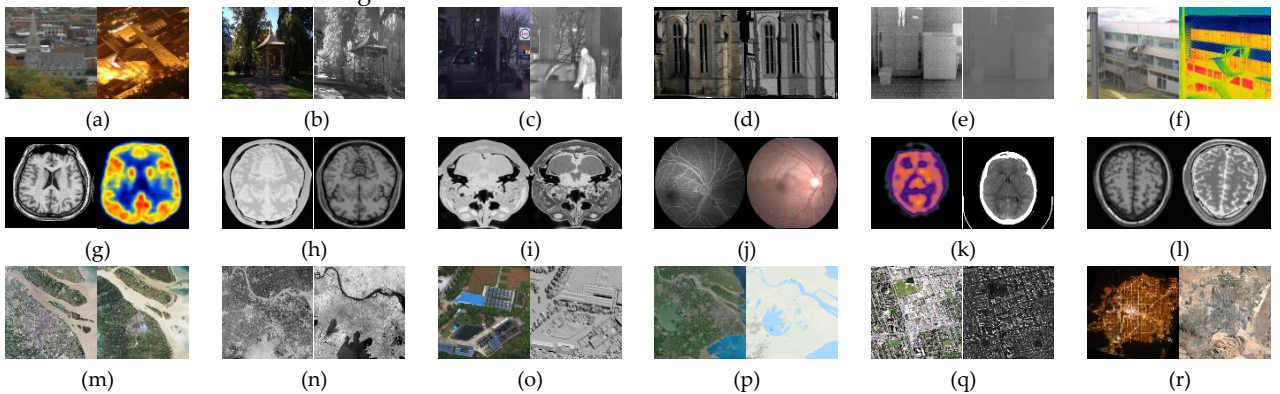

Fig.8. Typical multimodal image pairs from dataset 1 (computer vision), dataset 2 (medicine) and dateset 3 (remote sensing). (a) day-night; (b) visible-near infrared; (c) visible-infrared; (d) visible-LiDAR intersity;(e) visbile-LiDAR depth;(f) visible-thermal;(g) MRI-PET;(h) PD-T1;(i) PD-T2; (j)retina- retina (different angiography); (k)SPECT-CT; (l)T1-T2; (m)optical-optical (different season); (n)optical-infrared; (o)optical-LiDAR depth; (p)optical-map; (q)optical-SAR; (r)night-day.

TABLE 2
EXPERIMENTAL DATASETS

| Dataset | | day night | visible near-infrared | visible infrared | visible LiDAR-intensity | visible LiDAR-depth | visible thermal |
|---|---|---|---|---|---|---|---|
| Dataset 1 in Computer Vision, 38 pairs | Image pair | day night | visible near-infrared | visible infrared | visible LiDAR-intensity | visible LiDAR-depth | visible thermal |
| | Number | 4 | 7 | 11 | 6 | 4 | 6 |
| | Resolution | 263 × 198 ~ 1119 × 1465 pixels | | | | | |
| Dataset 2 in Medicine, 64 pairs | Image pair | MRI PECT | PD T1 | PD T2 | retina retina | SPECT CT | T1 T2 |
| | Number | 7 | 10 | 10 | 23 | 4 | 10 |
| | Resolution | 181 × 217 ~ 1280 × 960 pixels | | | | | |
| Dataset 3 in Remote Sensing, 62 pairs | Image pair | optical optical | optical infrared | optical LiDAR-depth | optical map | optical SAR | night day |
| | Number | 14 | 7 | 7 | 10 | 14 | 10 |
| | Resolution | 350 × 426 ~ 1001 × 1001 pixels | | | | | |

### 7.1 Parameter Study

We test the performance of NISR with respect to the scale of the image pyramid, $s$, the orientation number of the log-Gabor filter, $o$, the number of sub-regions inside a local image window, $n$, and the size of the image window for feature description, $l$. Notably, the sizes of the two image windows are set to $l$. We modify the tested parameter during experiments while keeping the other parameters unchanged. Besides, the maximum number of features of NISR is fixed at 1500. Table 3 provides the detailed experimental settings and the detailed results.

We vary $s$ from 2 to 6, and the matching performance reaches the best when $s = 4$. The number of matches will decrease and the *RMSE* will increase when s is smaller or larger than 4. Meanwhile, the performance improves with the increase of $s$ and $l$. In particular, when $s >= 2$ and $l >= 72$, the number of matches is larger than 829, and the *RMSE* is smaller than 1 pixel. Besides, we can see that *SR* is very sensitive to $n$ and robust against the other parameters. A larger $n$ helps to increase the matching results.

However, a larger $n$ will increase the dimensions of the feature vector, decreasing the efficiency. Based on the above analysis, we set $s = 4$, $o = 12$, $l = 72$, and $n = 6$ to give good results in most situations, and the settings are employed in the following experiments. Moreover, the maximum number of features of NISR is set to 5000.

TABLE 3
THE PERFORMANCE OF NISR UNDER DIFFERENT PARAMETER SETTINGS.

| Parameter setting | | | | NM ↑ | RMSE(pixels)↓ | SR(%)↑ |
|---|---|---|---|---|---|---|
| s | o | l | n | | | |
| 2 | 12 | 72 | 6 | 735 | 1.21 | 83.5 |
| 3 | 12 | 72 | 6 | 806 | 1.05 | 92.1 |
| 4 | 12 | 72 | 6 | 829 | 0.98 | 98.8 |
| 5 | 12 | 72 | 6 | 809 | 1.12 | 92.1 |
| 6 | 12 | 72 | 6 | 763 | 1.38 | 90.2 |
| 4 | 6 | 72 | 6 | 816 | 1.13 | 87.8 |
| 4 | 8 | 72 | 6 | 823 | 1.11 | 91.5 |
| 4 | 10 | 72 | 6 | 824 | 1.06 | 94.5 |
| 4 | 14 | 72 | 6 | 831 | 1.01 | 98.8 |
| 4 | 12 | 36 | 6 | 752 | 1.05 | 87.8 |
| 4 | 12 | 48 | 6 | 782 | 1.03 | 91.5 |
| 4 | 12 | 60 | 6 | 806 | 0.98 | 93.9 |
| 4 | 12 | 84 | 6 | 832 | 0.94 | 98.8 |
| 4 | 12 | 72 | 2 | 804 | 1.31 | 86.6 |
| 4 | 12 | 72 | 3 | 810 | 1.03 | 90.2 |
| 4 | 12 | 72 | 4 | 820 | 1.01 | 95.1 |
| 4 | 12 | 72 | 5 | 820 | 0.98 | 97.6 |

### 7.2 Performance with respect to scale and rotation change

In this section, we verify the robustness of NISR against scale change and image rotation. In terms of image rotation, we manually change the rotation angle between the matching images from 0° to 360° for three image pairs: a pair of visible-thermal images, a pair of Retina images with different angiography, and a pair of optical-map images. The quantitative experimental results measured by *NM* for NISR are presented in Fig. 9.

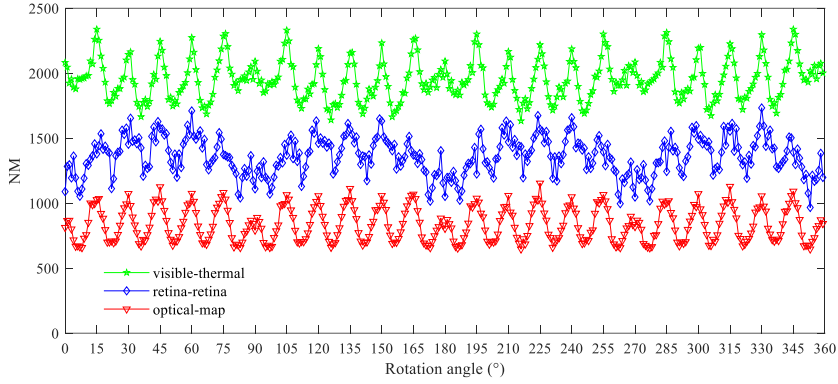

Fig.9. The changing curve of number of matches with different angles of image rotation for a VIS-INT image pair, a Retina-Retina image pair and an optical-map image pair.

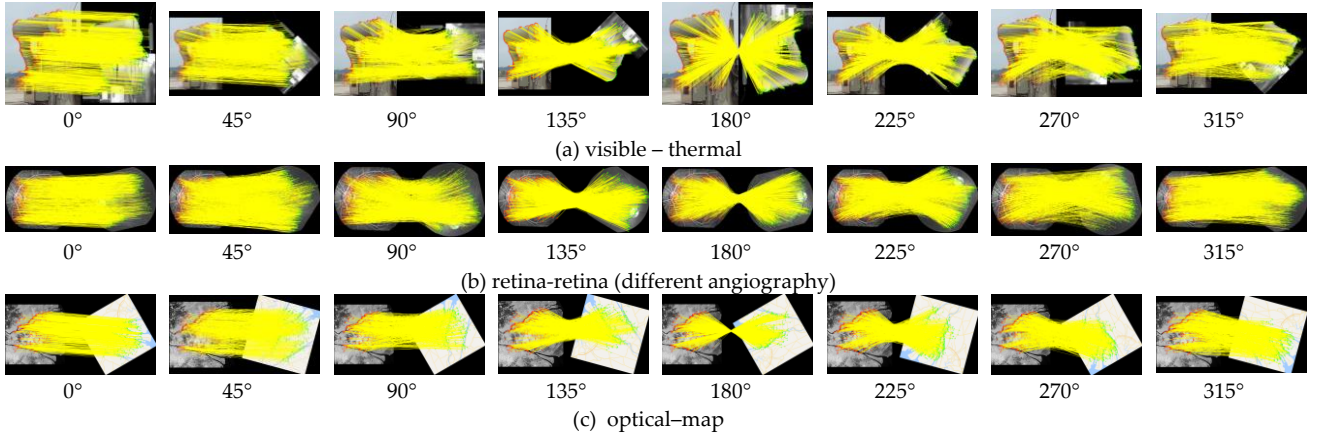

Fig.10. Some typical visualization results of Figure 9. The angle below an image represents the rotation angle between the image pair.

We can see that NISR can always keep a relatively high and stable performance even though there is a small range of period fluctuance. The periodicity includes a larger period of 90° and a smaller period of 15°. The large period is caused by image simulation, which is independent from our method. Specifically, the information of the simulated images is exact at every rotation of 90°, which will produce a fluctuance period of 90°. While the smaller periodicity is caused by the number of orientations $o$ of the log-Gabor filter. In the experiments, we set $o$ as 12, and the angle difference between two adjacent filters is 15°. When the rotation angle is $i \times 7.5°$, $i = 1,2,3,\cdots,48$, the index value of a pixel has a similar probability to its corresponding two adjacent index values, bringing the largest confusion and leading to the lowest similarity of the two orientation index maps. Meanwhile, the two orientation index maps will be the most similar at angles of $i \times 15°$, $i = 1,2,3\cdots,24$. Theoretically, the larger the value of $o$, the less affected by image rotation. Several typical visualization results at different rotations are displayed in Fig. 10.

In terms of scale invariance, we manually change the scale ratio between three image pairs, which are a pair of visible-infrared images, a pair of Retina images with different angiography, and a pair of optical-SAR images from 1:1 to 6:1, with their original size all 1000 ×1200 pixels. The quantitative experimental results are shown in Fig. 11.

We can see that $NM$ is proportional to image size and gradually decreases with the increase of scale ratio, demonstrating good scale invariance. In particular, when the scale ratio reaches 6:1, the width and height of the sensed image are less than 200 pixels, and there are still more than 180 matches, which meets the needs of most applications. The visualization results corresponding to Fig. 11 are displayed in Fig. 12.

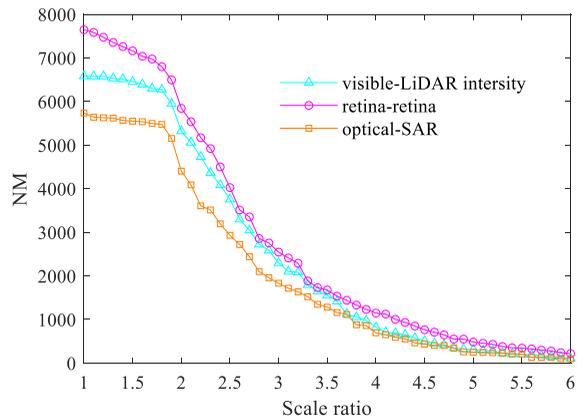

Fig.11. The changing curve of number of matches with different ratios of scale change.

## 7.3 Comparative qualitative evaluation

In this section, we select six representative image pairs of different categories for experiments. The visual comparison results of OS-SIFT, RIFT, CoFSM, LNIFT, and NISR are shown in Fig.13~Fig.15. For NISR, we show not only its final results but also the results of its initial matching stage, which are noted as NISR$_{ini}$.

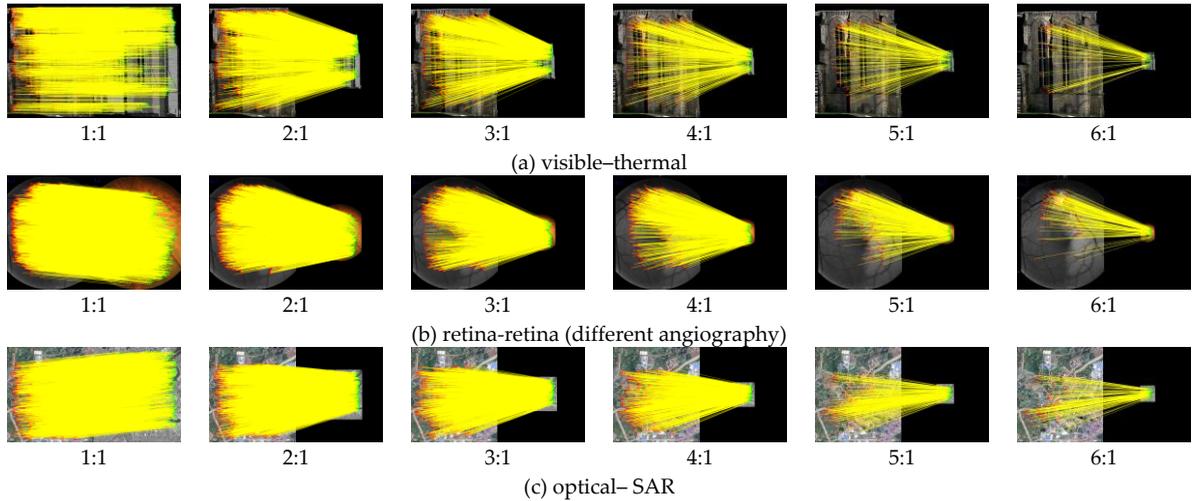

Fig.12. Some typical visualization results of Figure 11. The ratio below an image represents the scale difference.

We can see that OS-SIFT only matched 5 image pairs. RIFT exhibits good resistance to NID and successfully matched 13 of the 18 image pairs. However, RIFT failed to match the image pairs with scale differences and rotation, such as the pair of visible-infrared images. CoFSM is sensitive to the image modalities, matching most of the image pairs in dataset 1, half of the image pairs in dataset 2, and only the optical-infrared image pair in dataset 3. Even though LNIFT matched all 18 image pairs, the obtained matches are incorrect for at least half of the image pairs. In terms of our methods, NISR performed significantly better than all the other methods, and it is the only one that correctly matched all 18 image pairs. The feature matching stage of NISR (NISR$_{ini}$) is robust to critical NID and geometric differences, and the template matching stage can further improve the number of matches and matching accuracy. Therefore, NISR can be effectively applied to various multimodal image matching applications.

### 7.4 Comparative quantitative evaluation

The detailed quantitative results on all datasets are summarized in Table 4. We can see that NISR with/-out template matching achieved the best performance on all three metrics. The *NM* obtained by NISR was several or even dozens of times that of the best of the other methods, namely, CoFSM in datasets 1 and 3 and RIFT in dataset 2. For *RMSE*, we can see that NISR without template matching can keep a high accuracy, slightly worse than RIFT. However, after template matching, the accuracy improves to around 0.98 pixels on average, almost one time higher than RIFT. In terms of *SR*, NISR successfully matches almost all 164 image pairs except for two image pairs while the highest *SR* of the other methods is only 75.6%, proving the excellent robustness to various image modalities.

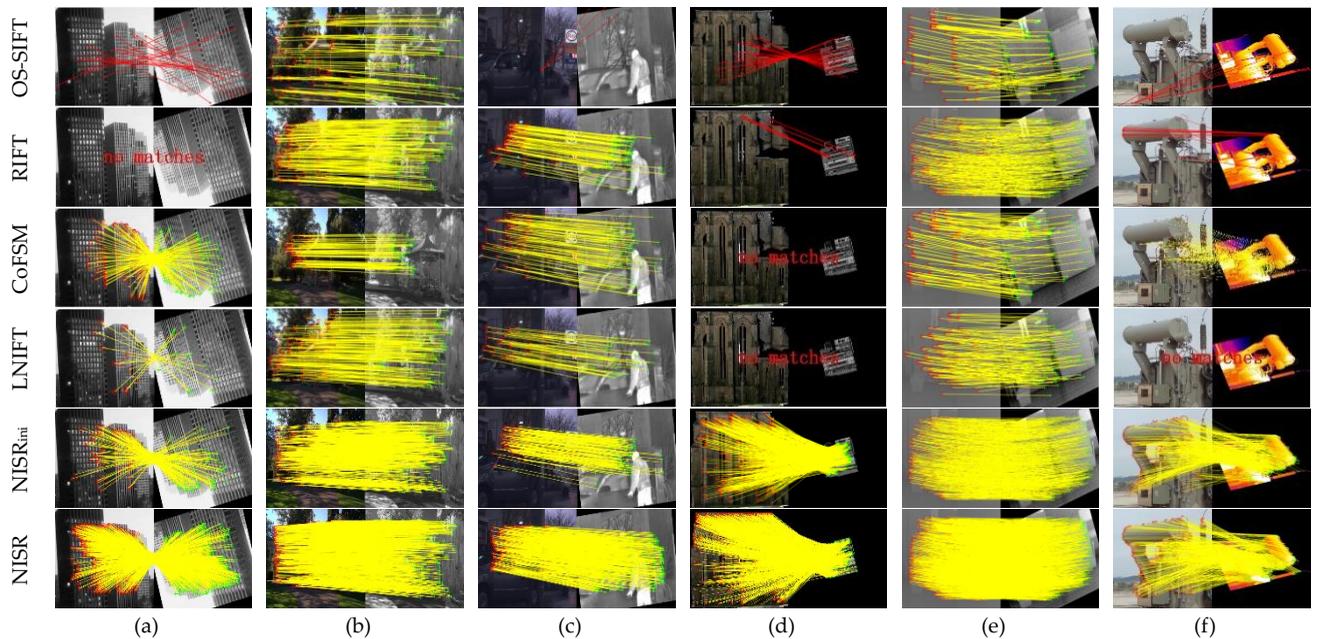

Fig. 13. The Qualitative comparison results of OS-SIFT, RIFT, CoFSM, LINIFT, NISR$_{ini}$. NISR on the six typical image pairs of dataset 1. (a) day-night; (b) visible-near infrared; (c) visible-infrared; (d) visible-LiDAR intersity;(e) visbile-LiDAR depth;(f) visible-thermal.

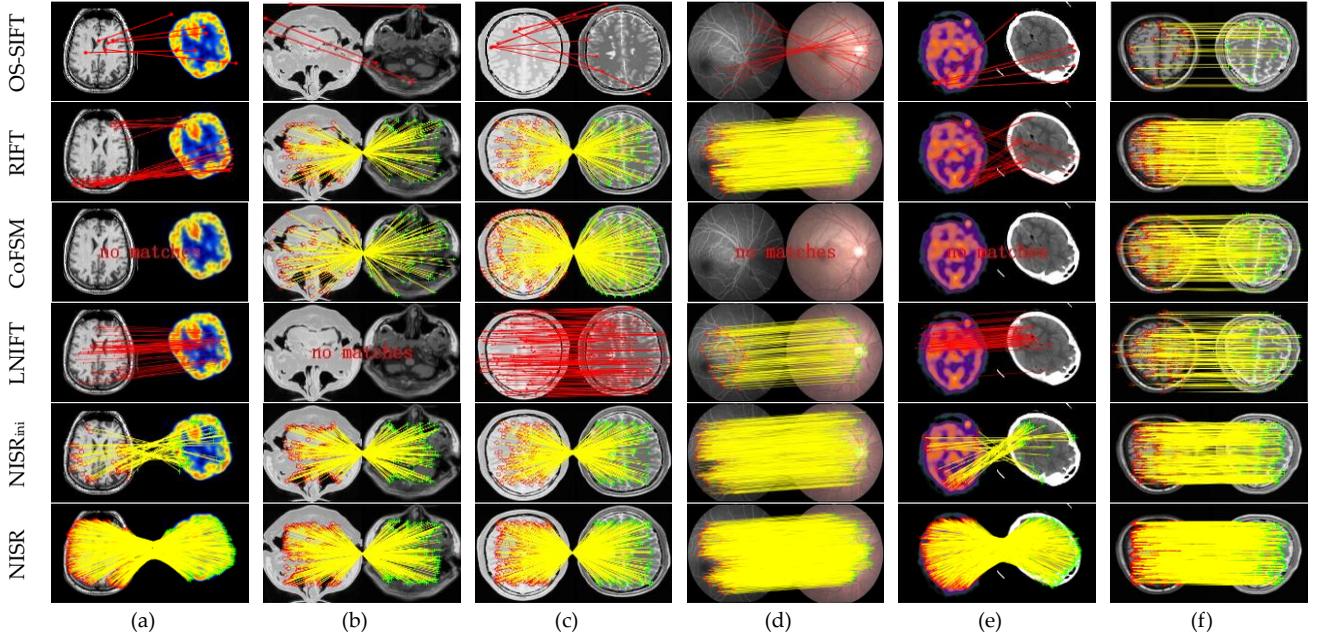

Fig. 14. The Qualitative comparison results of OS-SIFT, RIFT, CoFSM, LINIFT, NISR$_{ini}$. NISR on the six typical image pairs of dataset 2. (a) MRI-PET;(b) PD-T1;(c) PD-T2; (d) retina- retina (different angiography); (e)SPECT-CT; (f)T1-T2.

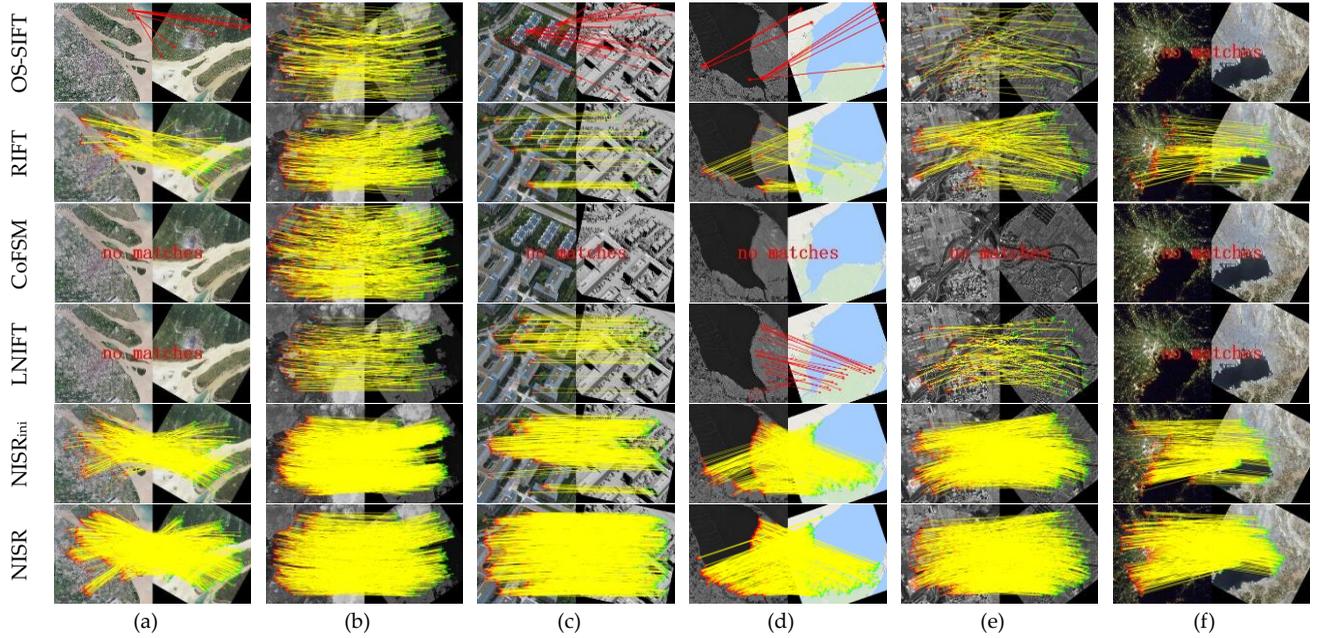

Fig. 15. The Qualitative comparison results of OS-SIFT, RIFT, CoFSM, LINIFT, NISR$_{ini}$. NISR on the six typical image pairs of dataset 3. (a)optical-optical (different season); (b)optical-infrared; (c)optical-LiDAR depth; (d)optical-map; (e)optical-SAR; (f)night-day.

## 8 IMAGE REGISTRATION

We apply NISR to the image registration and fusion tasks in this section. Taking the matches obtained from NISR, we calculate a transformation matrix, map each pixel of the sensed image to the coordinates of the reference image and create a corrected image with the bilinear interpolation. The performance is tested on six challenging multimodal image pairs: the visible-LiDAR Depth, visible-thermal, MRI-PECT, PD-T1, optical-map, and optical-SAR image pairs. The visualization results are presented in Fig. 16. We can see that the images are precisely registered and fused without edge break, ghosting and blurring, further proving the high accuracy and good distribution of the matches obtained from NISR.

## 9 CONCLUSION

In this paper, we present a robust multimodal image matching method, which can be effectively applied to various modalities of images. The whole process is based on multi-scale and multi-orientation log-Gabor filtering results. The filter makes the method naturally resistant to noise. Firstly, we detect distinctive and repeatable feature points on phrase congruency maps in an image pyramid. Secondly, we estimate the primary orientation by exploiting the index information from different image orienta-tions to make the method invariant to image rotation. Thirdly, we construct a template feature, fully applying the available filtering results and increasing the probability of successful matching. We demonstrate the benefits of our method on a large number of image datasets with various image modals.

TABLE 4
THE COMPARATIVE RESULTS ACROSS ALL DATASETS. VIS, NIR, IR, LI, LD, THERM, RETI, AND OPTI DEONTE VISIBLE, NEAR INFRARED, INFRARED, LIDAR INTENSITY, LIDAR DEPTH, THERMAL, RETINA, AND OPTICAL, RESPECTIVELY.

| | | NM ↑ | | | | | | RMSE (pixels) r ↓ | | | | | | SR (%) γ ↑ | | | | | |
|---|---|---|---|---|---|---|---|---|---|---|---|---|---|---|---|---|---|---|---|
| | | OS-SIFT | RIFT | CoFSM | LNIFT | NISR$_{ini}$ | NISR | OS-SIFT | RIFT | CoFSM | LNIFT | NISR$_{ini}$ | NISR | OS-SIFT | RIFT | CoFSM | LNIFT | NISR$_{ini}$ | NISR |
| Dataset 1 | day-night | 177 | 86 | 576 | 218 | **462** | 1864 | 1.86 | 3.11 | 1.99 | 2 | 2.18 | **1.02** | 75 | 50 | 100 | 75 | **100** | 100 |
| | VIS-NIR | 120 | 182 | 252 | 246 | **936** | 2111 | 2.3 | 2.67 | 2.78 | 3.92 | 2.61 | **1.08** | 100 | 71.4 | 85.7 | 28.5 | **100** | 100 |
| | VIS-IR | 17 | 81 | 99 | 170 | **351** | 1363 | 3.24 | 2.86 | 2.5 | 2.95 | 2.65 | **1.16** | 9.1 | 63.6 | 9.1 | 54.5 | **100** | 100 |
| | VIS-LI | 153 | 336 | 1448 | 235 | **2101** | 3427 | 1.65 | 1.43 | 1.52 | 1.99 | 1.76 | **0.81** | 33.3 | 83.3 | 83.3 | 50 | **100** | 100 |
| | VIS-LD | 87 | 448 | 435 | 158 | **1414** | 2752 | 2.69 | 1.96 | 1.33 | 2.96 | 1.62 | **0.81** | 50 | 75 | 50 | 75 | **100** | 100 |
| | VIS-Therm | 60 | 45 | 260 | 0 | **1773** | 2414 | 3.19 | 3.16 | 2.01 | 5 | 2.45 | **0.90** | 33.3 | 16.6 | 33.3 | 0 | **100** | 100 |
| Dataset 2 | MRI-PET | 0 | 0 | 0 | 0 | **70** | 640 | 5 | 5 | 5 | 5 | 4.42 | **2.06** | 0 | 0 | 0 | 0 | **100** | 100 |
| | PD-T1 | 29 | 80 | 164 | 175 | **165** | 281 | 1.77 | 2.16 | 2.84 | 2.82 | 1.56 | **0.69** | 80 | 90 | 90 | 70 | **100** | 100 |
| | PD-T2 | 26 | 84 | 182 | 225 | **177** | 286 | 2.46 | 1.87 | 2.36 | 2.83 | 1.37 | **0.69** | 90 | 90 | 100 | 80 | **100** | 100 |
| | Reti-Reti | 70 | 701 | 385 | 253 | **3067** | 3223 | 2.52 | 1.64 | 1.47 | 3.2 | 1.34 | **0.89** | 43.5 | 100 | 69.5 | 26.1 | **100** | 100 |
| | SPECT-CT | 0 | 0 | 0 | 46 | **62** | 565 | 5 | 5 | 5 | 4.4 | 4.39 | **1.63** | 0 | 0 | 0 | 25 | **100** | 100 |
| | T1-T2 | 36 | 120 | 184 | 338 | **237** | 411 | 1.59 | 1.15 | 1.49 | 2.17 | 1.02 | **0.64** | 90 | 90 | 100 | 50 | **100** | 100 |
| Dataset 3 | Opti-Opti | 109 | 251 | 406 | 73 | **704** | 1875 | 2 | 1.64 | 2.56 | 2.48 | 2.35 | **0.96** | 7.1 | 71.4 | 64.2 | 35.7 | **100** | 100 |
| | Opti-IR | 163 | 330 | 587 | 95 | **1321** | 2331 | 1.49 | 1.5 | 2.62 | 3.17 | 1.41 | **0.61** | 28.6 | **100** | 100 | 42.8 | 85.7 | 85.7 |
| | Opti-LD | 62 | 224 | 360 | 133 | **725** | 2475 | 1.5 | 1.13 | 3.14 | 3.63 | 2.19 | **0.81** | 14.3 | 85.7 | 28.5 | 14.2 | **100** | 100 |
| | Opti-map | 0 | 143 | 339 | 66 | **610** | 2237 | 5 | 2.23 | 1.88 | 3.56 | 2.28 | **0.91** | 0 | 90 | 30 | 50 | **100** | 100 |
| | Opti-SAR | 80 | 235 | 396 | 141 | **820** | 2424 | 1.76 | 1.9 | 3.05 | 3.33 | 2.66 | **1.21** | 7.1 | 78.5 | 28.5 | 50 | **100** | 100 |
| | night-day | 104 | 186 | 468 | 62 | **587** | 1926 | 2.47 | 3.36 | 3.07 | 3.03 | 2.81 | **1.16** | 10 | 70 | 40 | 40 | **90** | 90 |
| All | Average accuracy | 70 | 281 | 391 | 173 | **1012** | 1883 | 2.15 | 1.96 | 2.2 | 2.93 | 2.15 | **0.98** | 35.9 | 75.6 | 57.3 | 42.1 | **98.8** | 98.8 |

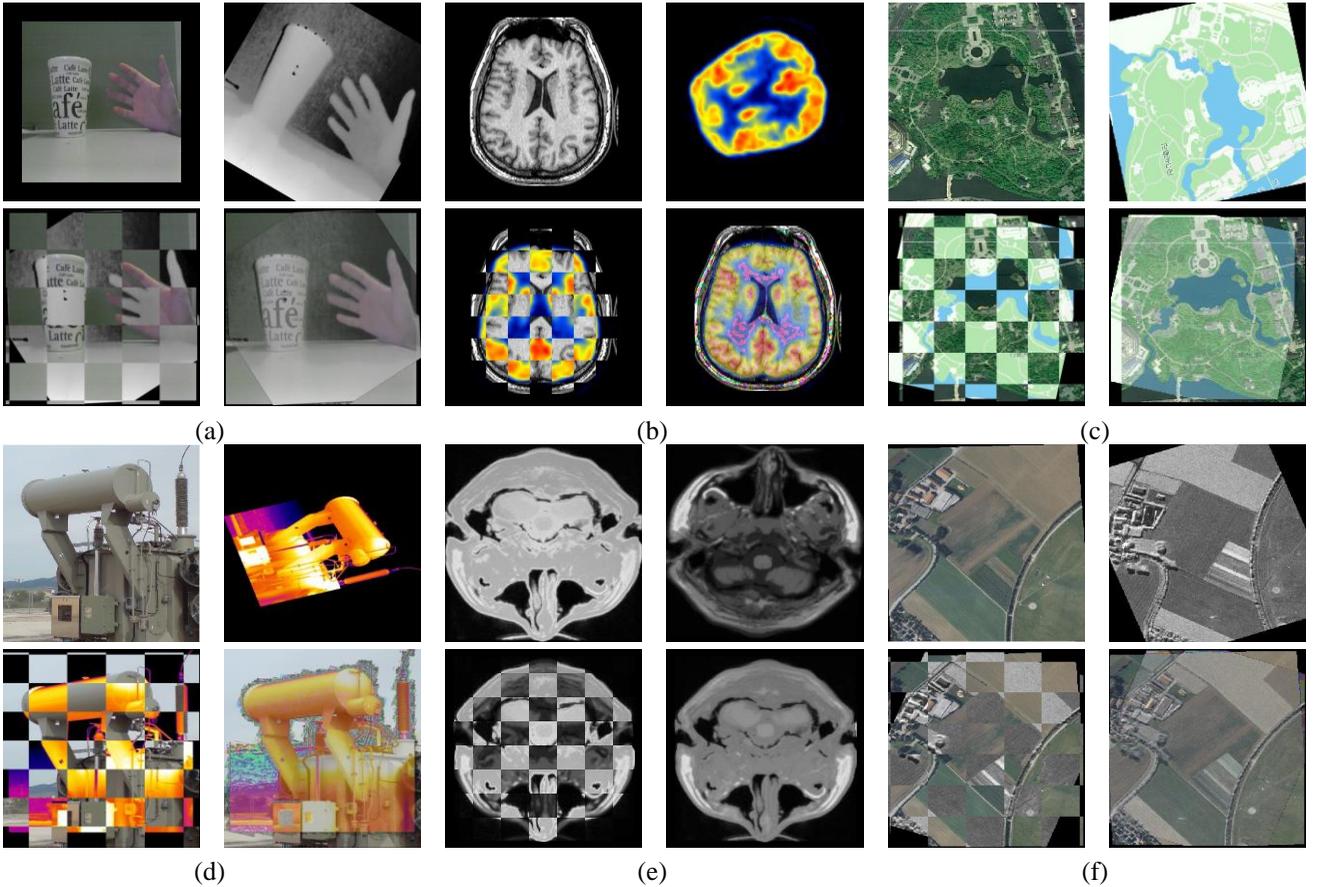

Fig.16. The image registration and fusion results of NISR on six multimodal image pairs. (a) visible-LiDAR depth; (b) MRI-PET; (c) optical-map; (d) visible-thermal; (e) PD-T1; (f) optical-SAR; For each subfigure (a)-(f), the two images in the first line are the original images, the images in left and right of the second line are the registration and fusion results, respectively.

In the future, we tend to apply convolutional neural networks to generate more accurate and robust orientation index maps, which will benefit exact primary orientation evaluation and feature descriptor construction. Besides, considering the high computation cost of log-Gabor filters, we will try other light weighted filters or approaches to extract the multi-scale and multi-orientation image information to improve the processing efficiency.

## ACKNOWLEDGMENT

The authors wish to thank Dr. Haibin Ai for his helpful suggestions. We also thank anonymous reviewers for their valuable feedback.

## REFERENCES

[1] D. Lowe, "Distinctive Image Features from Scale Invariant Key Points," *Int. J. Comput. Vis.*, vol. 60, no. 2, pp. 91- 110, 2004.

[2] Y. Ke and R. Sukthankar, "PCA-SIFT: A More Distinctive Representation for Local Image Descriptors," in *Proc. IEEE Conf. Comput. Vis. Pattern Recognit.*, 2004.

[3] J. Chen et al., "WLD: A robust local image descriptor," *IEEE Trans. Pattern Anal. Mach. Intell.*, vol. 32, no. 9, pp. 1705–1720, Sep. 2010.

[4] A. Sedaghat, M. Mokhtarzade, and H. Ebadi, "Uniform robust scale-invariant feature matching for optical remote sensing images," *IEEE Trans. Geosci. Remote Sens.*, vol. 49, no. 11, pp. 4516–4527, Nov. 2011.

[5] J. Ma, H. Zhou, J. Zhao, Y. Gao, J. Jiang, and J. Tian, "Robust feature matching for remote sensing image registration via locally linear transforming," *IEEE Trans. Geosci. Remote Sens.*, vol. 53, no. 12, pp. 6469–6481, 2015.

[6] B. Xu, L. Zhang, Y. Liu, H. Ai, B. Wang, Y. Sun, and Z. Fan, "Robust hierarchical structure from motion for large-scale unstructured image sets," *ISPRS J. Photogramm. Remote Sens.*, vol. 181, pp. 367–384, 2021.

[7] J. Engel, V. Koltun, and D. Cremers, "Direct sparse odometry," *IEEE Trans. Pattern Anal. Mach. Intell.*, 2017.

[8] J. Ma, X. Jiang, A. Fan, J. Jiang, and J. Yan, "Image matching from handcrafted to deep features: A survey," *Int. J. Comput. Vis.*, vol. 129, no. 1, pp. 23–79, 2021.

[9] X. Jiang, J. Ma, G. Xiao, Z. Shao, and X. Guo, "A review of multimodal image matching: Methods and applications," *Inf. Fusion.*, vol. 73, pp. 22–71, 2021.

[10] B. Zitoví and J. Flusser, "Image registration methods: A survey," *Image Vis. Comput.*, vol. 21, no. 11, pp. 977–1000, Oct. 2003.

[11] C. A. Aguilera, A. D. Sappa, and R. Toledo, "LGHD: A feature descriptor for matching across non-linear intensity variations," in *Proc. IEEE Int. Conf. Image Process.*, Sep. 2015, pp. 178–181.

[12] Y. Ye, J. Shan, L. Bruzzone, and L. Shen, "Robust registration of multimodal remote sensing images based on structural similarity," *IEEE Trans. Geosci. Remote Sens.*, vol. 55, no. 5, pp. 2941–2958, Mar. 2017.

[13] Y. Ye, L. Bruzzone, J. Shan, F. Bovolo, and Q. Zhu, "Fast and robust matching for multimodal remote sensing image registration," *IEEE Trans. Geosci. Remote Sens.*, vol. 57, no. 11, pp. 9059–9070, Nov. 2019.

[14] J. Li, Q. Hu, and M. Ai, "RIFT: Multi-modal image matching based on radiation-variation insensitive feature transform," *IEEE Trans. Image Process.*, vol. 29, pp. 3296–3310, 2020.

[15] J. Chen, J. Tian, N. Lee, J. Zheng, R. T. Smith, and A. F. Laine, "A partial intensity invariant feature descriptor for multimodal retinal image registration," *IEEE Trans. Biomed. Eng.*, vol. 57, no. 7, pp. 1707–1718, Jul. 2010.

[16] Y. Xiang, F. Wang, and H. You, "OS-SIFT: A robust SIFT-like algorithm for high-resolution optical-to-SAR image registration in suburban areas," *IEEE Trans. Geosci. Remote Sens.*, vol. 56, no. 6, pp. 3078–3090, Jun. 2018.

[17] Y. Yao, Y. Zhang, Y. Wan, X. Liu, X. Yan, and J. Li, "Multimodal remote sensing image matching considering co-occurrence filter," *IEEE Trans. Image Process.*, vol. 31, pp. 2584–2597, 2022.

[18] Z. Fan, Y. Liu, Y. Liu, L. Zhang, J. Zhang, Y. Sun, H. Ai, "3MRS: An Effective Coarse-to-Fine Matching Method for Multimodal Remote Sensing Imagery," *Remote Sens.*, vol. 14, no. 3, pp. 478, 2022.

[19] J. Li, W. Xu, P. Shi, Y. Zhang, and Q. Hu, "LNIFT: Locally normalized image for rotation invariant multimodal feature matching," *IEEE Trans. Geosci. Remote Sens.*, vol. 60, pp. 1–14, 2022.

[20] E. Rublee, V. Rabaud, K. Konolige, and G. Bradski, "ORB: An efficient alternative to SIFT or SURF," in *Proc. IEEE Int. Conf. Comput. Vis.*, Barcelona, Spain, 2011, pp. 2564–2571.

[21] N. Dalal and B. Triggs, "Histograms of oriented gradients for human detection," in *Proc. IEEE Comput. Soc. Conf. Comput. Vis. Pattern Recognit.*, vol. 1, Jun. 2005, pp. 886–893.

[22] F. Maes, A. Collignon, D. Vandermeulen, G. Marchal, and P. Suetens, "Multimodality image registration by maximization of mutual information," *IEEE Trans. Med. Imag.*, vol. 16, pp. 187–198, Apr. 1997.

[23] X. Liu, S. Chen, L. Zhuo, J. Li, and K. Huang, "Multi-sensor image registration by combining local self-similarity matching and mutual information," *Front. Earth Sci.*, vol. 12, no. 4, pp. 779–790, 2018.

[24] X. Xiong, Q. Xu, G. Jin, H. Zhang, and X. Gao, "Rank-based local selfsimilarity descriptor for optical-to-SAR image matching," *IEEE Geosci. Remote Sens. Lett.*, vol. 17, no. 10, pp. 1742–1746, 2020.

[25] Z. Fan, L. Zhang, Y. Liu, Q. Wang, and S. Zlatanova, "Exploiting high geopositioning accuracy of SAR data to obtain accurate geometric orientation of optical satellite images," *Remote Sens.*, vol. 13, no. 17, pp. 3535, 2021.

[26] B. Zhu, Y. Ye, L. Zhou, Z. Li, and G. Yin, "Robust registration of aerial images and LiDAR data using spatial constraints and Gabor structural features," *ISPRS J. Photogramm. Remote Sens.*, vol. 181, pp. 129–147, 2021.

[27] P. Kovesi, "Phase congruency detects corners and edges," in *Proc. Austral. Pattern Recognit. Soc. Conf. DICTA*, 2003, pp. 309–318.

[28] E. Rosten, R. Porter, and T. Drummond, "Faster and better: A machine learning approach to corner detection," *IEEE Trans. Pattern Anal. Mach. Intell.*, vol. 32, no. 1, pp. 105–119, Jan. 2010.

[29] D. J. Field, "Relations between the statistics of natural images and the response properties of cortical cells," *J. Opt. Soc. Amer. A*, vol. 4, no. 12, pp. 2379–2394, 1987.

[30] B. Horn, Robot Vision. Cambridge, MA, USA: MIT Press, 1986.

[31] Y. Wu, W. Ma, M. Gong, L. Su, and L. Jiao, "A novel point-matching algorithm based on fast sample consensus for image registration," *IEEE Geosci. Remote Sens. Lett.*, vol. 12, no. 1, pp. 43–47, Jan. 2015.

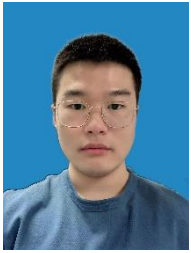

**Zhongli Fan** received his M.S. degree at Chinese Academy of Surveying and Mapping (CASM) and is currently pursuing a Ph.D. at Wuhan university, State Key Laboratory of Information Engineering in Surveying, Mapping and Remote Sensing. His research interest is in photogrammetry and computer vision problems, particularly image matching technologies with both photogrammetry and computer vision applications.

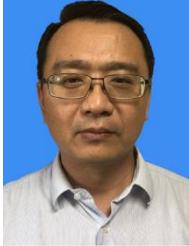

**Li Zhang** received the Ph.D. degree from Swiss Federal Institute of Technology in Zurich (ETH Zurich), Zurich, Swiss Confederation in 2005. He is currently a full-time professor at Chinses Academy of Surveying and Mapping (CASM), Beijing, China. His research interests include digital photogrammetry, computer vision, optical and SAR satellite image processing, and Multi-source Spatial Data Fusion. He has published over 100 papers at highly refereed coferences and journals including ISPRS/TGRS. His research has been generously supported by the National Key Technology R&D Program of China, National Natural Science Foundation of China and Sub-project of the 863 Program. He is a recipient of the Lufthansa Image Award.

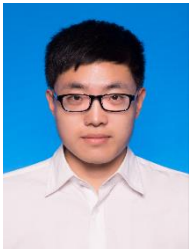

**Yuxuan Liu** received the Ph.D. degree of Engineering from Wuhan University, China, in 2020. He is currently an assistant professor in institute of Photogrammetry and Remote Sensing, Chinses Academy of Surveying and Mapping (CASM), Beijing, China. From 2018 to 2019, he was a visiting student in GRID lab, The University of New South Wales (UNSW), Sydney, Australia. His research interests include light field, photogrammetry and computer vision.